\documentclass[A4]{article}
\usepackage{graphicx}
\usepackage{xcolor}
\usepackage[hyphens]{url}
\usepackage{hyperref}
\usepackage[T1]{fontenc}
\usepackage{cite}
\usepackage{authblk}
\usepackage[margin=1.5in]{geometry}

\begin{document}
\title{Detecting corruption in single-bidder auctions via positive-unlabelled learning \thanks{Natalya Goryunova and Artem Baklanov are supported by the Russian Science Foundation grant 20-71-00034. Egor Ianovski is supported by the HSE University Basic Research Program}}
%
%

\author{Natalya Goryunova\textsuperscript{\rm 1}}
\author{Artem Baklanov\textsuperscript{\rm 2}}
\author{Egor Ianovski\textsuperscript{\rm 3}}
\affil{
    HSE University, Russian Federation \\
    \textsuperscript{\rm 1}ntl.goryunova@gmail.com, \textsuperscript{\rm 2}apbaklanov@hse.ru, \textsuperscript{\rm 3}george.ianovski@gmail.com
}
%
%
%
\maketitle              
\begin{abstract}
In research and policy-making guidelines, the single-bidder rate is a commonly used proxy of corruption in public procurement used but ipso facto this is not evidence of a corrupt auction, but an uncompetitive auction. And while an uncompetitive auction could arise due to a corrupt procurer attempting to conceal the transaction, but it could also be a result of geographic isolation, monopolist presence, or other structural factors. In this paper we use positive-unlabelled classification to attempt to separate public procurement auctions in the Russian Federation into auctions that are probably fair, and those that are suspicious.
\end{abstract}
\section{Introduction}

Public procurement is the process by which government entities purchase goods or services from the public sector. Given that a public official is spending public money on public interests, moral hazard is at play -- if a school director is given full freedom in how to purchase milk for a canteen, what is to stop him from buying the most expensive milk possible from a supplier that he happens to own stock in? As such strict regulations are needed to govern public procurement, typically specifying admissible auction formats and rules on advertisement and disclosure of tenders.

However, legislating a competitive and transparent procedure does not guarantee that the procedure will indeed be competitive and transparent. Corruption imposes a cost on society, and scholars seeking to understand this cost need to devise means to measure how prevalent corruption is \cite{Cai2013,Yakovlev,Igraham,Fazekas2020}. A common proxy of corruption in public procurement is the single-bidder rate -- the proportion of auctions that only attracted a single firm -- which is both prominent in the literature \cite{Klasnja2015,Charron,Wachs,Fazekas2013} and is used by the European Commission to assess the effectiveness of public procurement in EU member states \cite{SingleMarketScoreboard}.

The single-bidder rate in Russia is high. In the class of auctions studied in this paper 48\% attracted a single bidder; compared to the EU only Poland and Czechia fare worse with 51\% \cite{SingleMarketScoreboard}. However, while the problem of corruption in Russia is well known \cite{Transparency2019,AcChamber}, there are many reasons why the procurement process could be uncompetitive -- the country is vast and sparsely populated, and large sectors of the economy are dominated by monopolies and oligopolies; it is conceivable that in many parts of the country if an official wishes to purchase a product, one supplier is the best he can hope for. Learning to divide the two is important, as while addressing the problems of monopolies requires long-term structural change, corruption can be dealt with immediate regulatory action.

\subsection{Related work}

The idea of this work originated in a series of works on bid leakage in Russian procurement auctions \cite{Andreyanov18,Ivanov,Ivanov2019}. Bid leakage is a form of corruption in a first-price, sealed-bid auction where the procurer reveals the contents of other firms' bids to a favoured firm, allowing the favoured firm to submit a marginally lower bid at the end of the auction and take the contract for the highest price possible.

The work of \cite{Andreyanov18} is based on the assumption that in a fair auction the order of the bids should be independent of the winner; if it turns out that the last bidder is more likely to win than the others, there is reason to suspect bid-leakage has taken place. This was followed up by \cite{Ivanov2019,Ivanov} who relaxed the independence assumption, as they demonstrated that in a game theoretic model there are legitimate reasons for a serious competitor to delay bidding -- in particular, if a firm believes the procurer is corrupt and bid leakage could take place, by bidding near the deadline there would be no time to leak the bid to the favoured firm. Their approach was based on the DEDPUL positive-unlabelled classifier \cite{Dedpul}: auctions where the last bidder did not win were labelled as ``fair'' (even if bid leakage did take place, it was not successful), and the classifier separated the remaining auctions into ``fair'' or ``suspicious''. Their approach found signs of bid leakage in 9\% of auctions with three or more participants, and 16\% of auctions with two or more.

The single-bidder rate is a common proxy of corruption in the literature, but while the authors acknowledge that a single-bidder auction is not necessarily evidence of corruption, there has been no attempt to distinguish the two. To our knowledge, the closest is the works of \cite{Charron} and \cite{Fazekas2020}, who find some correlation between the single-bidder rate and features suggestive of corruption such as a short advertisement period or a subjective award criteria. We are not aware of any work that attempts to assign a posteriori probabilities to a single-bidder auction being corrupt.

\subsection{Our contribution}

We use the DEDPUL positive-unlabelled classifier \cite{Dedpul} to separate a dataset of auctions held in the Russian Federation in the years 2014--2018 into a class that is ``fair'' and ``suspicious'', based on a selection of features indicative of corruption. This approach labels just over half (53.86\%) of single-bidder auctions as ``suspicious''. The distribution of posterior probabilities reveals a cluster of auctions with the posterior probability of being labelled as ``suspicious'' close to 1. A decision tree for this cluster reveals two patterns that resemble the ``one-day firm'' mode of corruption -- a firm being created on paper to snap up government contracts rather than do legitimate business -- and a third pattern that could equally describe a legitimate monopolist as a corrupt relationship.

\section{Methodology}
\subsection{Positive-unlabelled  learning}
In this paper, we employ the method of positive-unlabelled (PU) learning that, similar to general binary classification, provides a classifier that can separate positive and negative instances based on the features but with less information available. Namely, the training (labelled) data constitute only a fraction of the positives; labels for negative examples are not provided.

A PU dataset can be mathematically abstracted as a collection of $(x, y, s)$ with $x$ a vector of features of an instance, $y$ the class and $s$ a binary variable  indicating whether the triplet was labelled. The true class $y$ may not be known. By definition of a PU learning problem, every instance with $s = 1$ is labelled as positive: $Pr(y = 1 | s = 1) = 1.$

Following the general PU learning setting, we assume that the data $\mathbf{x}$ is  an independent and identically distributed sample from $f$, the probability density function (pdf) that we want to estimate, such that 
$$ \mathbf{x} \sim f(x) = f_u(x)= \alpha f_{+} (x) + (1- \alpha) f_{-} (x) ;$$
here $\alpha$ stands for the prior of positive class (i.e., $Pr(y = 1)$) and $f_+, f_-, f_u$ are the pdfs of the positive, negative, and unlabelled examples, respectively.

Unfortunately, the true value of $\alpha$ is unidentifiable \cite{Dedpul}. Namely, even under precise knowledge of $f_u$ and $f_+,$ estimation of $\alpha$ is an ill-posed problem since any $\hat{\alpha}$ such that $f_u (\cdot) \geq \hat{\alpha} f_+ (\cdot)$ is a valid guess. Thus, following \cite{Dedpul} we only compute $\alpha^*,$ the upper bound for the valid estimates of true $\alpha,$
\begin{equation}
\label{alpha_star}
    \alpha^* = \inf_{x} \frac{f_u(x)}{f_+(x)}, 
\end{equation}
hence $ \alpha \in [0, \alpha^*]$. 

We make the following assumption, a rather strong one but common for PU learning (see \cite{Elkan} and \cite[Definition 1]{Bekker}):  the labelled examples were Selected Completely At Random (SCAR) independently from their attributes. Thus, we treat the probability of absence of corruption to be independent from the attributes of auctions and equal to the estimate of prior $\alpha^*.$ 
This assumption delivers nice properties for classification problems (see \cite{Elkan}):
\begin{itemize}
    \item The probability of an instance being labelled is proportional to the probability of an instance being positive.
    \item Non-traditional classifiers (classifiers that treat unlabelled instances as negative) preserve the ranking order. I.e., if we only wish to rank instances with respect to the
chance that they belong to positive class, then non-traditional classifiers rank instances similarly to estimation of $f$ obtained by a traditional probabilistic classifier. 
\end{itemize}

A robust estimation of $\alpha^*$ can be a very complex task due to challenges of finding the best fit for multidimensional empirical distributions and the $\inf$ operation in (\ref{alpha_star}). 
To tackle this issue, we employ a state-of-the-art PU learning algorithm (DEDPUL \cite{Dedpul}) that uses multiple regularisation techniques. 
At the first step,  a non-traditional classifier is trained. In general,  this can be any classifier separating positive from unlabelled. At this step, using cross-validation, we applied CatBoost
\cite{Prokhorenkova2019},  an algorithm based on gradient boosting of decision trees that achieves state-of-the-art performance.
At the second step, DEDPUL corrects the bias caused by treating all unlabelled instances as negatives, taking care of the challenges we mentioned.


\subsection{The data}

Public procurement in Russia is governed by Federal Law No. 44-FZ \cite{FederalLaw44}, which specifies admissible procurement formats and requires that the data be publicly available on the official website (\url{https://zakupki.gov.ru/}). We focus on the ``requests for quotation'' format which is a first-price, sealed-bid auction for low value transactions (the maximum reserve price is 500,000 roubles, approximately 4,880 GBP). These are frequent auctions with an objective award criterion (i.e., the contract is awarded to the lowest bid, with no considerations of quality or reputation) and are thus amenable to machine learning techniques.

The dataset used in this paper was extracted by \cite{Ivanov}. It covers the years 2014--2018 and consists of 3,081,719 bids from 1,372,307 auctions and 363,009 firms. An observation in the dataset is a bid and is labelled by the identification of the procurer, firm, auction, and region; the reserve price of the auction and the actual bid of the firm; the start and end date of the auction; the date the bid was actually placed.\footnote{All the data and code used is available on request.}

After preliminary processing we removed 10.6\% of the bids. About 3\% removed due to errors in the data, consisting in one or more of:
\begin{enumerate}
    \item Missing values in the bid description.
    \item The start date of the auction being later than the end date.
    \item The bid amount being less than zero or higher than the reserve price.
    \item The reserve price being higher than the maximum allowed price of 500,000 roubles.
\end{enumerate}
The rest were removed for one of three reasons:
\begin{enumerate}
\item The auction took place in Baikonur, which is administered by Russia but is part of Kazakhstan.
\item The reserve price was under 3,440 roubles (lowest 0.5\%).
\item The firm placing the bid appears once in the dataset.
\end{enumerate}
Baikonur was excluded for its peculiar status. The minimum reserve price of 3,440 (about 30 GBP) is an ad hoc approach to remove potential data errors -- a price of 0 or 1 rouble should probably be classed as an error, but it is not clear where to draw the line, so we opted to drop the bottom 0.5\%.

Firms that bid once were removed because one of our features is the length of time the firm is active in the system. This is a potential issue since our dataset covers the years 2014--2018 and could capture a firm that was active before this period and stopped in 2014, or a firm that began activity in 2018. We do not wish to misidentify an established firm with a long history of bids that ceased operations in 2014 with a firm that only placed one bid, ever. 

This left us with 2,787,136 bids. Every bid consists of four identifiers and five values. The ranges of these values are summarised in Table~\ref{tab:ids} and Table~\ref{tab:variables}.

\begin{table}[h!]
\centering
\caption{\label{tab:ids} Categorical variables}
\begin{tabular}{|c|c|c|}
\hline
\textbf{Variable} & \textbf{Description} & \textbf{Number of values}   \\
\hline
$\mathtt{procurer\_id}$ & Identification of the procurer & 43,311\\
\hline
$\mathtt{firm\_id}$ & Identification of the firm & 255,650\\
\hline
$\mathtt{auction\_id}$ & Identification of the auction & 1,358,369 \\
\hline
$\mathtt{region\_id}$ & The auction location (RF subject) & 85\\
\hline
\end{tabular}
\end{table} 

\begin{table}[h!]
\begin{center}
\caption{\label{tab:variables} Numeric variables}
\begin{tabular}{|c|c|c|c|c|c|}
\hline
\textbf{Variable} & \textbf{Description}& \textbf{Min}& \textbf{Median} & \textbf{Max}  \\
\hline
$\mathtt{reserve\_price}$ & Reserve price set by procurer (roubles)  & 3,440 & 134,637  & 500,000	\\
\hline
$\mathtt{price}$& Bid price set by firm (roubles)  & 0.01 & 106,500 & 500,000\\
\hline
$\mathtt{start\_date}$ & Start date of the auction   & 28.01.14 &  & 26.03.18\\
\hline
$\mathtt{end\_date}$ & End date of the auction & 31.01.14 &  & 30.03.18\\
\hline
$\mathtt{date}$ & Time of bid &  29.01.14 & & 26.03.18\\
\hline
\end{tabular}
\end{center}
\end{table} 

The single-bidder rate was high, and increasing over the time period.

\begin{center}
   \begin{tabular}{|c|c|c|c|c|c|}
   \hline
    Year &  2014&2015&2016&2017&2018\\
     \hline
     Single-bidder rate&0.4&0.47&0.51&0.52&0.51\\
     \hline
\end{tabular} 
\end{center}

\subsection{Feature engineering}

We train the classifier on the features in Table~\ref{tab: features}. These were chosen to reflect potential signs of corruption. The intuition behind the features is as follows:

\begin{table}[h]
\begin{center}
\caption{\label{tab: features} Features engineered for the model}
\begin{tabular}{|p{6.2cm}|c|c|c|c|c|}
\hline
\textbf{Variable} & \textbf{ID} & \textbf{Type} & \textbf{Min}& \textbf{Median}& \textbf{Max}  \\
\hline
Is the auction a single-bid auction?& $\mathtt{single}$  & Binary & 0 & 0 & 1 \\
\hline
Time from bid to the end date (seconds)& $\mathtt{bid.date}$ & Int & 0 & 72 000 & 783 840 \\
\hline
Ratio of bid to reserve price & $\mathtt{bid.price}$ & Float & 0 & 0.90 & 1 \\
\hline
Has the firm dealt with the procurer before? & $\mathtt{con.met}$ & Binary & 0 & 0 & 1 \\
\hline
Ratio of firm's victories with procurer to total victories & $\mathtt{con.win}$ & Float & 0& 0.06 &1 \\
\hline
How many auctions did the firm bid in? & $\mathtt{sel.num}$ & Int & 2 & 27 & 8319 \\
\hline
How long the firm is active in the data (days) & $\mathtt{sel.period}$ & Int & 0 & 980 & 1 498 \\
\hline
Ratio of reserve price to maximum (500,000) & $\mathtt{au.reserve}$ & Float & 0 & 0.27 & 1 \\
\hline
Auction duration (days) & $\mathtt{au.duration}$ & Int & 0 & 7 & 26 \\
\hline
Is auction in Moscow or Moscow Oblast? & $\mathtt{au.moscow}$ & Binary & 0 & 0 & 1 \\
\hline

Ratio of unique winners to number of auctions held by procurer & $\mathtt{buy.unique}$ & Float & 0.02 & 0.5 & 1 \\
\hline
\end{tabular}
\end{center}
\end{table} 

\begin{itemize}
    \item $\mathtt{bid.date}$: If a firm knows in advance there will be no competition, they have no need to delay their bid.
    \item $\mathtt{bid.price}$: If a firm knows there will be no competition, they will ask for the highest possible price.
    \item  $\mathtt{con.met}$: If a firm has a corrupt relationship with a procurer, they are likely to have met before.
    \item $\mathtt{con.win}$: If a firm wins a disproportionate number of its tenders from a single procurer, the question arises why this firm is unable to win an auction anywhere else.
    \item $\mathtt{sel.num}$: A monopolist would likely participate in a vast amount of auctions; a firm owned by the procurer is likely smaller.
    \item $\mathtt{sel.period}$: A monopolist would be an established firm.
    \item $\mathtt{au.reserve}$: If you are out to fleece the taxpayer, why ask for less than the maximum price?
    \item $\mathtt{au.duration}$: A short auction is less likely to be noticed.
    \item $\mathtt{au.moscow}$: The ``geographic isolation'' argument does not apply to the heart of the Russian economy.
\end{itemize}

The main feature, $\mathtt{single}$, is whether or not the auction attracted a single firm. We label auctions with more than one participant ($\mathtt{single}=0$) as ``fair'' and the remaining auctions ($\mathtt{single}=1$) are the unlabelled set that the classifier will attempt to separate into ``fair'' and ``suspicious''. As shown in Table~\ref{tab: comparison}, these classes are behave differently with respect to our features.

\begin{table}[h!]
\begin{center}
\caption{\label{tab: comparison} Statistics of single-bidder and competitive auctions}
\begin{tabular}{|c|c|c|c|c|c|c|}
\hline
& \multicolumn{2}{c|}{\textbf{Mean}} & \multicolumn{2}{|c|}{\textbf{Median}} & \multicolumn{2}{|c|}{\textbf{Std. Dev.}} \\
\cline{2-7}
\raisebox{1.5ex}[0cm][0cm]{\textbf{ID}}
& $\mathtt{single}=1$ & $\mathtt{single}=0$ & $\mathtt{single}=1$ & $\mathtt{single}=0$ & $\mathtt{single}=1$ & $\mathtt{single}=0$  \\
\hline
$\mathtt{bid.date}$ & 146 910 & 133 815 & 76 800 & 71 220 & 172 297 & 167  765\\
\hline
$\mathtt{bid.price}$& 0.94 & 0.81 & 0.99 & 0.86 & 0.12 & 0.19  \\
\hline
$\mathtt{con.met}$ & 0.58 & 0.43 & 1 & 0 & 0.49 & 0.50  \\
\hline
$\mathtt{con.win}$ & 0.42 & 0.20 & 0.29 & 0.02 & 0.37 & 0.32  \\
\hline
$\mathtt{sel.num}$ & 288 & 181 & 22 & 29 & 1 070 & 683 \\
\hline
$\mathtt{sel.period}$ & 927 & 855 & 1064 & 947 & 422 & 439 \\
\hline
$\mathtt{au.reserve}$ & 0.30 & 0.38 & 0.20 & 0.30 & 0.29 & 0.31\\
\hline
$\mathtt{au.duration}$ & 7.83 & 8.26 & 7 & 7 & 2.83 & 2.94 \\
\hline
$\mathtt{au.moscow}$ & 0.06 & 0.14 & 0 & 0  & 0.25 & 0.35 \\
\hline
$\mathtt{buy.unique}$ & 0.49 & 0.54 & 0.48 & 0.52 & 0.18 & 0.19 \\
\hline
\end{tabular}
\end{center}
\end{table} 

\section{Results}

The DEDPUL classifier was trained on the features in \autoref{tab: features}. The positive class is all auctions with $\mathtt{single}=0$. The unlabelled auctions ($\mathtt{single}=1$) were separated into positive and negative. The algorithm found an $\alpha^*$ of 46.14\%. Thus $1-\alpha^*$ -- the probability of a single-bidder auction being labelled negative -- is 53.86\% (13\% of all bids).

The distribution of posterior probabilities has an interesting cluster near~1. 26\% of all bids in single-bidder auctions (about 170,000) have a posterior probability of being negative higher than 0.96. These bids differ strongly from those in multi-bidder auctions, but it is too early to conclude that they are corrupt -- it is conceivable that an auction in a monopolistic market, where both the procurer and the monopolist know the auction will only attract a single bidder, will look very different from multi-bidder auctions. For example, the monopolist might always bid the reserve price, knowing the bid will be unchallenged, and an honest procurer might counter this by using a lower reserve price than they would in a competitive market.

\begin{figure}[h!]
\center{\includegraphics[width= 10cm, height = 6cm ]{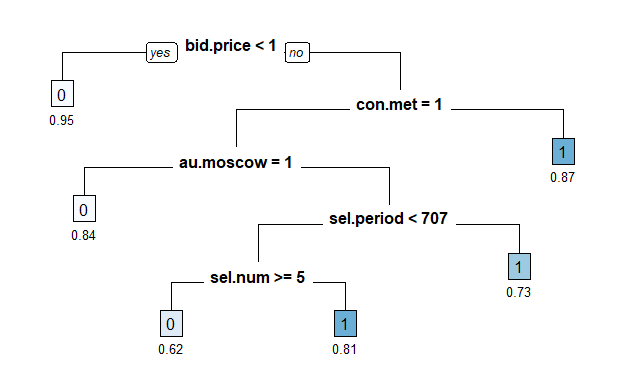}}
\caption{Decision tree that detects highly suspicious auctions (class 1) within the set of all single-bid auctions. \label{ris:tree_unfair}}
\end{figure}

To understand the nature of this cluster better we fit a classification tree, depicted in \autoref{ris:tree_unfair}, that separates two classes of single-bid auctions: the one consisting of all highly suspicious actions (class 1) with posterior probability of being negative higher than 0.96 and the complementing class that includes the remaining auctions  (class 0). The tree shows high  overall accuracy equal to 0.9. 

There are three paths to an auction being placed in the cluster by the tree:

\begin{enumerate}
    \item If bid price equals to reserve price and the firm has not met the procurer before.
    \item If bid price equals to reserve price, the firm has met the procurer, the auction is not in Moscow or Moscow Oblast, and the firm has been active in the system for more than 707 days.
    \item If bid price equals to reserve price, the firm has met the procurer, the auction is not in Moscow or Moscow Oblast, and the firm has been active in the system for less than 707 days and participated in less than 5 auctions.
\end{enumerate}

The fact that the firms bid the reserve price exactly is a nearly a prerequisite for an auction being in the cluster. This demonstrates that a firm, for whatever reason, knows that it will be the sole bidder, and does not even attempt to compete. But as we argued this is not necessarily corruption, since a monopolist would have no reason to compete either.

The first path, then, is interesting as this is a firm that expects no competition but has not dealt with the procurer before. A monopolist would likely have had dealings with all counterparties in its area; this path seems to better resemble a one-day firm created for the sole task of snapping up a contract. The third path, representing a short-lived firm that did not participate in many auctions, also resembles this form of corruption; such a firm certainly does not resemble a monopolist.

There is little we can say about the second path -- an established firm that dealt with the procurer before. This could be a monopolist, or a firm with an established corrupt relationship with the procurer.

It is curious that both the second and third path require that the auction does not take place in Moscow or Moscow Oblast. It is not obvious why this is the case -- ceteris paribus, we would consider a single-bidder auction in a highly competitive region like Moscow to be more such suspicious than one elsewhere. Perhaps due to greater policing, procurers are more careful to mask signs of corruption, and single-bidder auctions there more closely resemble the multi-bidder case.

\section{Discussion}

Our main result challenges the common assumption in the literature and policy-making that the single-bidder rate can serve as a good proxy of corruption in public procurement. Using state-of-the-art semi-supervised learning algorithm, we demonstrate that  multi-bidder and single-bidder auctions may belong to the same latent class of competitive auctions. Using PU learning we obtained an $\alpha^*$ -- the upper bound on the probability of a single-bidder auction belonging to the same category as multi-bidder auctions -- of 46.14\%. In other words, almost half of all single-bidder auctions are very similar to multi-bidder auctions, and could very well be fair. Conversely, this means that at least half are different, and could represent corruption. By ranking the single-bidder auctions based on their posterior probability to belong to the suspicious class we identified a cluster of auctions with very high posterior probabilities -- over 0.96. We built a decision tree for auctions to belong to this class, and find that at least two of the patterns identified by the tree fall resemble a form of corruption -- a one-day firm. 

We acknowledge that though the above threshold of 0.96 for highly suspicious single-bid auctions is to some extent arbitrary, the ranking of single-bid auctions' posterior probabilities of being corrupt is not (by the SCAR assumption). Thus, as an immediate implication for the state regulator, examination of the top-ranked auctions should be prioritised and become routine. 





The positive-unlabelled method used in this paper relies on two strong assumptions that should be relaxed in future works.
First, we assumed that all multi-bidder auctions are fair. This is clearly not the case: not only will there be cases where the procurer attempted, but failed, to restrict participation to a single firm, there could also be the situation where the procurer intentionally held a multi-bidder auction -- perhaps registering several one-day firms owned by himself to create the illusion of competition.\footnote{Compare with the two strategies a cartel may use to implement bid rotation: the other firms could refrain from bidding in an auction a certain cartel member is supposed to win, or they could place intentionally non-competitive bids to fool the regulators.} A natural future direction is then to try learning with noisy labels \cite{Northcutt2020} -- the single-bidder auctions are labelled as ``suspicious'', the multi-bidder as ``fair'', but both categories are assumed to contain mislabelled elements.

The second one is the SCAR assumption required by DEDPUL that treats the probability of corruption to be independent from the attributes of auctions and equal to $1-\alpha^*$, one minus the estimated prior probability of the positive class. This assumption can be relaxed to the Selected At Random (SAR) assumption (see \cite[Definition 2]{Bekker} and \cite{Bekker2019}) when the probability for choosing positive examples to be labelled is conditional on its features.

\Urlmuskip=0mu plus 1mu\relax
\bibliographystyle{splncs04}
\bibliography{mybibliography}

\begin{thebibliography}{10}
\providecommand{\url}[1]{\texttt{#1}}
\providecommand{\urlprefix}{URL }
\providecommand{\doi}[1]{https://doi.org/#1}

\bibitem{AcChamber}
{Accounts Chamber of the Russian Federation}: {Report on results of the
  analytical event «Monitoring of public and corporate procurement development
  in Russian Federation in 2018»}.
  \url{https://ach.gov.ru/promo/goszakupki-2018/index.html} (2018), accessed:
  2021-01-18

\bibitem{Andreyanov18}
Andreyanov, P., Davidson, A., Korovkin, V.: Detecting auctioneer corruption:
  Evidence from {R}ussian procurement auctions  (07 2018)

\bibitem{Bekker}
Bekker, J., Davis, J.: Learning from positive and unlabeled data: a survey.
  Machine Learning  \textbf{109},  719--760 (2020)

\bibitem{Bekker2019}
Bekker, J., Robberechts, P., Davis, J.: Beyond the selected completely at
  random assumption for learning from positive and unlabeled data (2019)

\bibitem{Cai2013}
Cai, H., Henderson, J.V., Zhang, Q.: China's land market auctions: evidence of
  corruption? The RAND Journal of Economics  \textbf{44}(3),  488--521 (2013)

\bibitem{Charron}
Charron, N., Dahlström, C., Lapuente, V., Fazekas, M.: Careers, connections,
  and corruption risks: Investigating the impact of bureaucratic meritocracy on
  public procurement processes. The Journal of Politics  \textbf{79} (10 2016).
  \doi{10.1086/687209}

\bibitem{Elkan}
Elkan, C., Noto, K.: Learning classifiers from only positive and unlabeled
  data. Proceedings of the 14th ACM SIGKDD international conference on
  Knowledge discovery and data mining p. 213–220 (2008)

\bibitem{SingleMarketScoreboard}
{European Commission}: Performance per policy area: Public procurement.
  \url{https://ec.europa.eu/internal_market/scoreboard/performance_per_policy_area/public_procurement/index_en.htm}
  (2019), accessed: 2021-01-22

\bibitem{Fazekas2013}
Fazekas, M., János, T., King, L.: Anatomy of grand corruption: A composite
  corruption risk index based on objective data. Tech. rep. (11 2013).
  \doi{10.2139/ssrn.2331980}

\bibitem{Fazekas2020}
Fazekas, M., Kocsis, G.: Uncovering high-level corruption: Cross-national
  objective corruption risk indicators using public procurement data. British
  Journal of Political Science  \textbf{50}(1),  155–164 (2020).
  \doi{10.1017/S0007123417000461}

\bibitem{Igraham}
Ingraham, A.: A test for collusion between a bidder and an auctioneer in
  sealed-bid auctions. Contributions in Economic Analysis \& Policy (4) (2005)

\bibitem{Dedpul}
Ivanov, D.: Dedpul: Method for mixture proportion estimation and
  positive-unlabeled classification based on density estimation. arXiv preprint
  arXiv:1902.06965  (2019)

\bibitem{Ivanov2019}
Ivanov, D., Nesterov, A.: Identifying bid leakage in procurement auctions:
  Machine learning approach. In: Proceedings of the 2019 ACM Conference on
  Economics and Computation. p. 69–70. EC '19 (2019).
  \doi{10.1145/3328526.3329642}

\bibitem{Ivanov}
Ivanov, D.I., Nesterov, A.S.: Stealed-bid auctions: Detecting bid leakage via
  semi-supervised learning (2020)

\bibitem{Klasnja2015}
Klasnja, M.: Corruption and the incumbency disadvantage: Theory and evidence.
  The Journal of Politics  \textbf{77} (08 2015). \doi{10.1086/682913}

\bibitem{Northcutt2020}
Northcutt, C.G., Jiang, L., Chuang, I.L.: Confident learning: Estimating
  uncertainty in dataset labels (2020)

\bibitem{Prokhorenkova2019}
Prokhorenkova, L., Gusev, G., Vorobev, A., Dorogush, A.V., Gulin, A.: Catboost:
  unbiased boosting with categorical features (2019)

\bibitem{FederalLaw44}
{Single Information System in the Sphere of Procurement}: {Federal Law No.
  44-FZ of 1 January 2014 "On the contract system in state and municipal
  procurement of goods, works and services"}.
  \url{https://zakupki.gov.ru/epz/main/public/download/downloadDocument.html?id=33991}
  (2014), accessed: 2021-01-18

\bibitem{Transparency2019}
{Transparency International}: Transparency international corruption perceptions
  index. \url{https://www.transparency.org/en/cpi} (2019), accessed: 2021-01-18

\bibitem{Wachs}
Wachs, J., Fazekas, M., Kertész, J.: Corruption risk in contracting markets: a
  network science perspective. International Journal of Data Science and
  Analytics pp. 1--16 (01 2020). \doi{10.1007/s41060-019-00204-1}

\bibitem{Yakovlev}
Yakovlev, A., et~al.: Incentives for repeated contracts in public sector:
  empirical study of gasoline procurement in russia. International Journal of
  Procurement Management (9),  2640 -- 2647 (2016)

\end{thebibliography}

\end{document}